\documentclass[runningheads]{llncs}
\usepackage{graphicx}
\usepackage{amsmath,amssymb} 
\usepackage{color}
\newcommand{\norm}[1]{\left\lVert#1\right\rVert}

\begin{document}
\title{Learnable Pooling Methods \\ for Video Classification} 

\titlerunning{Learnable Pooling Methods for Video Classification}
%
\author{
Sebastian Kmiec \and
Juhan Bae \and
Ruijian An}
%
\authorrunning{S. Kmiec and J. Bae and R. An}
%

\institute{University of Toronto, Toronto, Canada \\
\email{\{sebastian.kmiec, juhan.bae, ruijian.an\}@mail.utoronto.ca}}
\maketitle              
\begin{abstract}
We introduce modifications to state-of-the-art approaches to aggregating local video descriptors by using attention mechanisms and function approximations. Rather than using ensembles of existing architectures, we provide an insight on creating new architectures. We demonstrate our solutions in the "The 2nd YouTube-8M Video Understanding Challenge", by using frame-level video and audio descriptors. We obtain testing accuracy similar to the state of the art, while meeting budget constraints, and touch upon strategies to improve the state of the art. Model implementations are available in \newline https://github.com/pomonam/LearnablePoolingMethods.

\end{abstract}
\section{Introduction}
The problem of summarizing local descriptors is highly investigated and encompasses many domains in machine learning such as image retrieval. The goal of aggregating local descriptors is to construct a single global descriptor that encodes useful information.   
Although progress exists in the context of local video descriptor aggregation \cite{netvlad}, \cite{learnablepooling}, few works adequately provide solutions for differentiable descriptor aggregation, such as NetBoW \cite{learnablepooling} and SMK \cite{radenovic2018revisiting}, \cite{tolias2016image}. 

Aggregation of local descriptors extends to the task of video classification. Many existing models focus on learning temporal relations within a video. In particular, recurrent neural networks (RNN) with the help of Long Short Term Memory (LSTM) or gated rectified units (GRU) can capture the long-term temporal patterns in between frames. In this paper, however, we question the usefulness of learning temporal relationships in video classification. This is also largely motivated by the work of NetVLAD \cite{netvlad} and attention clusters \cite{attentionagg}, producing a dominant result compared to recurrent methods \cite{xie2017rethinking}, \cite{bian2017revisiting}.

Despite the success of NetVLAD for the task of video understanding \cite{learnablepooling}, many design choices are left unexplained. For example, the individual contribution of cluster center residuals to the global representation is unclear. The model also exhibits several weaknesses. For instance, the global representation generated by NetVLAD is projected to a significantly lower dimensional space for classification, which we suspect is difficult for a single layer to manipulate. To accommodate these issues, inspired by \cite{faemb}, \cite{attention}, we propose a model capable of learning inter-feature relationships before and after the NetVLAD layer.

\section{Related Work}
In this section, we outline previous works that heavily influenced the creation of our learnable pooling architectures
\subsection{Attention Mechanisms} \label{attmech}
Our work is largely inspired by \cite{attentionagg}, where for a given set of local descriptors, an attention representation is created via a simple weighted sum of local descriptors. These weights are computed as a function of the local descriptors to exclusively obtain information from useful descriptors in the attention representation. This attention representation is termed an attention cluster, and multiple clusters are concatenated to form a final global representation. A novel shifting operation, as in Equation \eqref{attentionaggeq}, enables each attention cluster to diverge from each other, while keeping scale invariance. Below is the output of a single attention cluster; \textbf{X} is a matrix of local descriptors, \textbf{a} is a vector of attention weights, computed via a simple two layer, feed-forward network.

\begin{align}
  \psi_{k}(X) &= \frac{\alpha \cdot \textbf{a}\textbf{X} + \beta}{\sqrt{N} \norm{\alpha \cdot \textbf{a}\textbf{X} + \beta}_{2}} \label{attentionaggeq} 
\end{align}

Although the above mechanism has the capacity to learn the training data, the model heavily overfits, and fails to generalize well. We hypothesize that as you are computing a weighted sum of input descriptors, slight changes in the distribution of inputs will have a large negative impact on performance. Intuitively, to avoid this problem, internal embeddings should be summed instead, and these internal embeddings should be a function of the inputs, to improve generalization.

Following the above logic is the work by A Vaswani et al \cite{attention}, whom achieved state-of-the-art performance for machine translation tasks, using a novel attention mechanism termed Transformer. In the context of their work, local embeddings are projected to query, key and value spaces. The similarity of keys (\textbf{K}) to queries (\textbf{Q}) is then used to provide weights to the internal embedding vectors (as in Equation \eqref{attentiontrans}) that are passed to feed forward networks. Both \cite{attentionagg} and \cite{attention} use dot product attention for attention-based representations. However, attention clusters intend to provide a straight-forward global representation, whereas a Transformer creates a new attention-based encoding of equal dimensionality. Our use of Transformers is covered in detail within Section \ref{arch}.

\begin{align}
  \text{Attention}(Q, K, V ) = \text{softmax}(\frac{\textbf{Q}\textbf{K}^{T}}{\alpha})\textbf{V} \label{attentiontrans} 
\end{align}

\subsection{Differentiable Pooling} \label{diffpool}
The work of \cite{learnablepooling} provides a useful baseline for a video classification model, as it provides the highest known accuracy of any single architecture, as of the "Google Cloud \& YouTube-8M Video Understanding Challenge". The authors utilized NetVLAD \cite{netvlad}, along with a novel network block that is comparable to residual blocks \cite{residual}, known as gating. Gating has the effect of re-weighting the importance of features detected, and decisions made based on these features.

VLAD itself summarizes a set of local descriptors by presenting these descriptors via a distribution \cite{vlad}. This distribution is encoded in the sum of distances to cluster centres. Specifically, looking at Equation \eqref{vladeq}, $a_{j}$ refers to the cluster soft similarity of descriptor $x_{k}$ to the jth cluster center, introduced by NetVLAD to avoid non-differentiable hard assignments. Here, $\boldsymbol{x}_{k} - \boldsymbol{c}_{j}$ refers to a single distance of a local descriptor to a cluster centre. This is a sound technique that is used in many areas \cite{girdhar2017actionvlad}, \cite{zhu2018attention}, while achieving state of the art performance in the benchmarks such as \cite{retrievalbenchmark}. 

\begin{align}
  VLAD(i, j) &= \sum_{k=1}^{N} a_{j}(x_{k}) \left(x_{k}(i)-c_{j}(i)\right)\; \label{vladeq}
\end{align}

Despite the success and prominence of VLAD, numerous design choices in \cite{learnablepooling} are left unexplained. For instance, the cluster similarities, $a_{j}$, are computed via a simple linear transformation, where each local video descriptor is compared via dot product with a key per cluster centre, followed by a softmax layer. It is a straightforward idea to consider multiple keys per cluster centre, or to use multiple temporally close local descriptors per cluster similarity prediction.

Further on the notion of design choices, another weakness arises in the architecture after the NetVLAD block. Aside from the use of gating, all outputs of the NetVLAD module are simply squeezed and/or projected to a low dimensional space for classification, this is too demanding of a task for a single layer to perform optimally. A final criticism is that this model does not attempt to leverage inter-feature relationships prior to the use of the NetVLAD module.

\subsection{Regularized Aggregation} \label{regagg}
Following the success of VLAD is the work of \cite{temb}. The authors split the problem of local descriptor pooling into problems of local descriptor embedding and aggregation. We specifically utilize the ideas for local descriptor embeddings.  The work of \cite{temb} creates a new embedding technique by focusing on overcoming pitfalls when using NetVLAD, by L2 normalizing the distances from cluster centers (residuals) before summing to help give an equal contribution to each residual. These embeddings are known as Triangulation embeddings, or T-embeddings. Further, the authors suggest whitening the residuals by removing a bias and de-correlating the residuals per cluster center, for a given local descriptor, before summing. 

We do utilize the above two ideas. However, the suggestion of using democratic weights before summing \cite{temb} is avoided. The use of democratic weights is intended to give each local descriptor an equal contribution to the similarity of the class they belong to. Nevertheless, it is not clear how to make an easily parallelizable implementation of the Sinkhorn scaling algorithm to obtain a solution for these weights. The use of weights is described in Equation \eqref{temb1}, where $\boldsymbol{X}$ is the set of local descriptors belonging to a given class, and $\boldsymbol{\phi}_{i}$ is an embedding per local descriptor, as displayed in Equation \eqref{temb2}. In Equation \eqref{temb3}, $\boldsymbol{\Sigma}$ is the covariance matrix of \textbf{R(X)}, where \textbf{R(X)} is a random variable representing $\boldsymbol{R}(x_{i})$ in the set of \textbf{X}, per class. A single $\boldsymbol{R}(x_{i})$ is the concatenation of normalized residuals to K cluster centers for a single local descriptor, as in \eqref{temb3}.

\begin{align}
  \boldsymbol{\psi}(\boldsymbol{X})       &= \sum_{i=1}^{N} \lambda_{i} \boldsymbol{\phi}_{i} \label{temb1}\\
  \boldsymbol{\phi}_{i}(\boldsymbol{x}_{i})   &= \boldsymbol{\Sigma}^{-1/2} (\boldsymbol{R}(\boldsymbol{x}_{i}) - E[\boldsymbol{R}(\boldsymbol{X})]) \label{temb2} \\
  \boldsymbol{R(x_{i})}     &= [\boldsymbol{r}_{1}(\boldsymbol{x}_{i}), \dots, \boldsymbol{r}_{K}(\boldsymbol{x}_{i})] \label{temb3}
\end{align}

\subsection{Function Approximations}
The work of \cite{faemb} builds upon \cite{localtangent}, which intends to provide an encoding per image, for the sake of image classification, using a weighted sum of local tangents at anchor points. Despite the dissimilarity of use cases for VLAD and tangent encoding, the authors of \cite{faemb} provide a mathematical formulation, to describe how given a certain set of cluster similarity weights, tangent encoding is a generalization of VLAD.

As tangent encoding is a technique to linearly approximate a high dimensional function \cite{faemb}, the authors naturally extend VLAD to a second order approximation to obtain a unique local descriptor embedding, while incorporating ideas from other methods such as \cite{temb} for aggregation. An embedding per local descriptor is described in Equation \eqref{faemb1}, where $\boldsymbol{\phi}_{i}(x_{i})$ is the concatenation of three vectors, per cluster center j. The $a_{j}(\boldsymbol{x}_{i})$ can be thought of as cluster similarities, as in Equation \eqref{vladeq}. 

What is newly introduced to VLAD is the flattened vector of $a_{j}(\boldsymbol{x}_{i}) \cdot F(\boldsymbol{v}_{i, j} \boldsymbol{v}_{i,j}^{T})$ (in this equation $F(\cdot)$ is the flattening operation), this provides the second order approximation of our hypothetical function, derived from a Taylor expansion \cite{faemb}. In addition, objective functions are provided to help regularize the learning of weights, so as to ensure a valid function approximation, this is further discussed in Section \ref{experimental}.

\begin{align}
  \boldsymbol{\phi}_{i}(\boldsymbol{x}_{i})   &=  [a_{j}(\boldsymbol{x}_{i});\;  a_{j}(\boldsymbol{x}_{i}) \cdot \boldsymbol{v}_{i, j}; \; a_{j}(\boldsymbol{x}_{i}) \cdot F(\boldsymbol{v}_{i, j} \boldsymbol{v}_{i,j}^{T})]_{j=1}^{K} \label{faemb1} \\
  \boldsymbol{v}_{i, j} &= (\boldsymbol{x}_{i} - \boldsymbol{c}_{j})
\end{align}

\section{Learnable Pooling Architectures} \label{arch}
In this section, we describe the architecture of our proposed learnable, pooling methods, for the purpose of video classification. As inputs to our pooling methods, we have audio and video features already extracted at the frame level per second of video, from the "YouTube-8M Dataset" \cite{8mbenchmark}. Our pooling methods aggregate all local descriptors per frame into a single global representation that describes a video, with possible post-processing afterwards. The final global video descriptor is then passed to a Mixture-of-Experts (MoE) \cite{moe} for classification, where probabilities are output across possible video labels.

\subsection{Attention Enhanced NetVLAD} \label{modelone}
\begin{figure}[!htb]
\centering
\includegraphics[width=\textwidth]{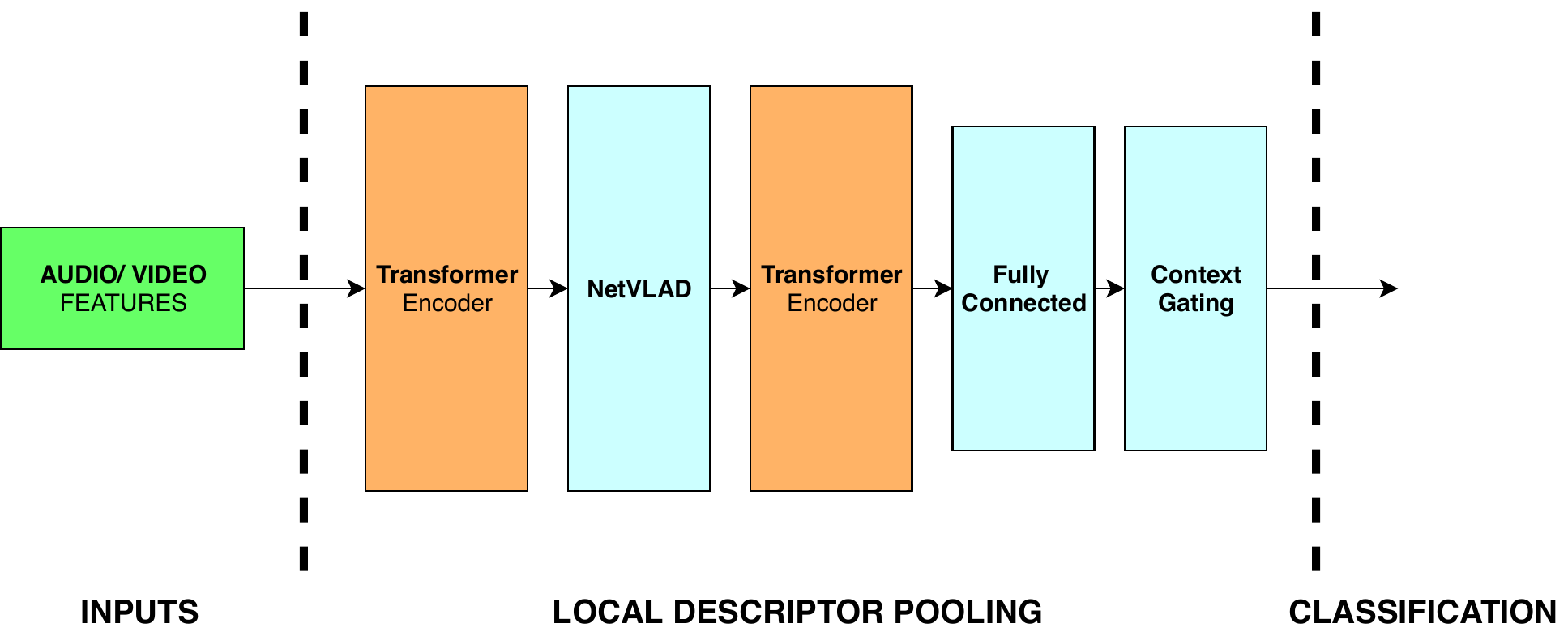}
\caption{A block diagram of an attention enhanced NetVLAD model. A modified Transformer Encoder \cite{attention} is placed before and after the NetVLAD module.}
\label{netvladatten1}
\end{figure}

Our first approach is to use a transformer encoder before and after a NetVLAD module, as in Figure \ref{netvladatten1}. Our local descriptor pooling is based largely on the work of \cite{learnablepooling}. As motivated in Section \ref{diffpool}, NetVLAD with context gating is the current (completely differentiable) state of the art for video pooling, as per benchmarks \cite{retrievalbenchmark}, and the result of the "Google Cloud \& YouTube-8M Video Understanding Challenge". Similarly, we have already motivated the use of Transformers in Section \ref{attmech}.

The first block is a mapping of $f_{1}: {\rm I\!R}^{N \times F} \rightarrow {\rm I\!R}^{N \times F}$, where N is the number of frame features sampled, and F is the feature size. The first transformer operates on the level of frames. As we uniformly sample frames per video as input, information relating to the relative positions of local descriptors inherently exists within the attention mechanism.

The second block is a mapping $f_{2}: {\rm I\!R}^{C \times F} \rightarrow {\rm I\!R}^{C \times F}$, where C is the number of NetVLAD cluster centers. The second block operates on the level of clusters. Our belief is that using separate query/key/value projection per cluster (as per the Transformer encoder), should be easier to learn than a single fully connected layer, that must perform the function of attention, decision making and dimensionality reduction all at once. We effectively spread the responsibility of these crucial functions across multiple layers, increasing the capacity of the initial work of \cite{learnablepooling}.

The Transformer Encoder in Figure \ref{netvladatten1} refers to the work of \cite{attention}, with batch normalization added to inner layers within the Multi-Head Attention block, to make learning a simpler process.

\subsection{NetVLAD with Attention Based Cluster Similarities} \label{modeltwo}

\begin{figure}[!htb]
\centering
\includegraphics[width=\textwidth]{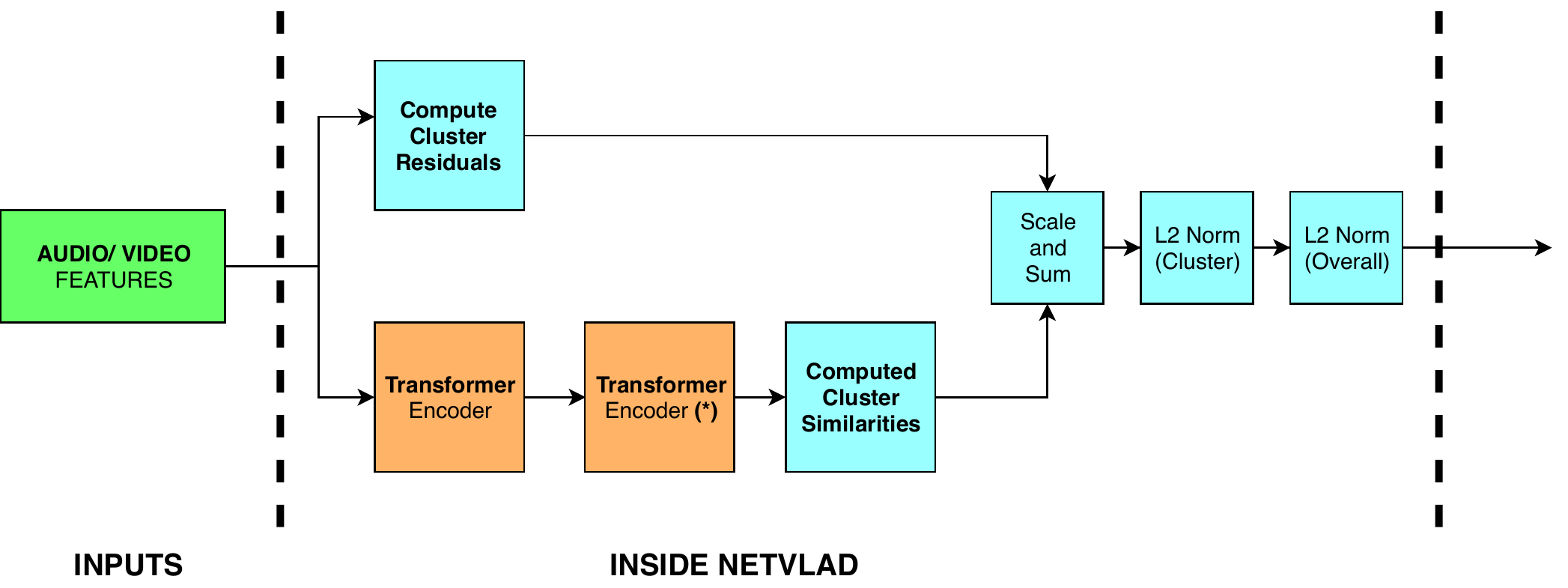}
\caption{A block diagram of a modified NetVLAD module. A pair of modified Transformer Encoders \cite{attention} is used to compute cluster similarities.}
\label{netvladatten2}
\end{figure}

Along with the aforementioned Transformer based model, we propose the following model, using modified NetVLAD modules. As in Figure \ref{netvladatten2}, the first Transformer encoder is a mapping of $g_{1}: {\rm I\!R}^{N \times F} \rightarrow {\rm I\!R}^{N \times F}$, whereas the second encoder is modified to be a mapping from $g_{2}: {\rm I\!R}^{N \times F} \rightarrow {\rm I\!R}^{C \times F}$.

Similar to Section \ref{modelone}, we argue that determining cluster similarities per local descriptor via a simple dot product with one key per cluster, followed by a softmax, is too demanding of a task (although it is simple to learn). Furthermore, cluster similarity is a task well suited for a Transformer-based attention mechanism, as the goal is simply to remember what input descriptors are highly correlated to (or similar) to which cluster centers, by using keys that are dependent on the inputs themselves. In addition, by using a Transformer, our prediction of input descriptor similarity to a cluster now receives information from other input descriptors, further improving the capacity of the initial work of \cite{learnablepooling}.

The Transformer Encoder (*) in Figure \ref{netvladatten2} is similar to the Transformer encoder discussed in Section \ref{modelone}, however, the final feed forward layer projects to a dimension of size C instead of F, and hence, the final residual connection is removed.

\subsection{Regularized Function Approximation Approach} \label{experimental}
Given our heavy reliance on NetVLAD \cite{netvlad} in our previous two models, our final model attempts to address issues that lie within NetVLAD and adaptations. The work of \cite{temb} already provides useful suggestions regarding possible pitfalls when using NetVLAD, by L2 normalizing the distances from cluster centers (residuals) before summing, as well as whitening these residuals. Unfortunately, we do not make use of democratic weights for aggregation of intermediate T-embeddings, as proposed in \cite{temb}, as even simplified versions of the Sinkhorn scaling algorithm (with assumptions made) are slow to train with.

\begin{figure}[!htb]
\centering
\includegraphics[width=\textwidth]{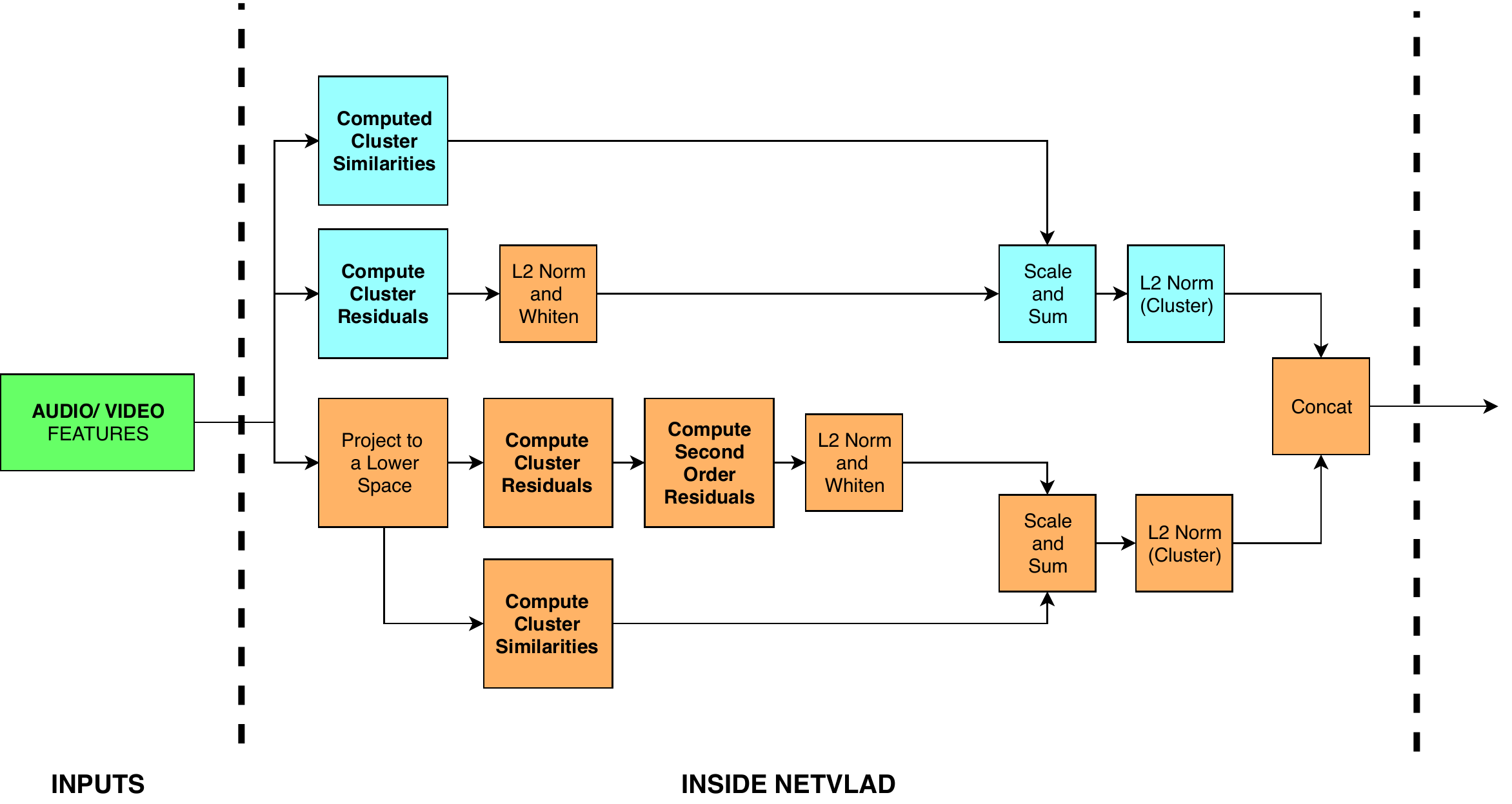}
\caption{A block diagram of the second order, function approximation based model. In addition to the usual NetVLAD implementation, we add residual normalization and whitening, along with second order terms based on projected input features.}
\label{netvlad3}
\end{figure}

To account for the discriminative power lost by missing democratic weights, we unite the works of \cite{temb}, \cite{faemb} and \cite{localtangent}, as illustrated in Figure \ref{netvlad3}. By adding second order terms, we now introduce more useful information in our global representation by adding multiplicative residual terms. To the point, these multiplicative terms cannot be computed by a simple linear transformation that follows our global representation, as is the case in \cite{learnablepooling}.

Notice that for the second order terms, we first project the inputs to a lower space, as the second order cluster residuals are elements of ${\rm I\!R}^{B \times N \times C \times F \times F}$, where B is the batch size, N is the number of local descriptors per video, C is the number of clusters and F is the input feature size. This is too large to fit even on multiple high-end commercial GPUs, given a feature size of 1024 (video features). Also note that as we perform a simple linear projection, as well as separate cluster centre and similarity computations for these squeezed features, it is expected that performance is to be lost.

Furthermore, we do not utilize the suggested regularizer terms within cost functions, as in \cite{faemb}. The reason being that regularization requires tuning in order to provide a valid contribution. Given the large amount of time required to train this model, we avoid additional regularization terms. Although regularizer terms such as in \cite{faemb}, or in \cite{clusterregularization} (for the sake of cluster center sparsity), may be necessary to overcome generalization issues discussed in Section \ref{exp}.

\section{Experiments} \label{exp}
Herein, we provide implementation details for experiments performed on models found in Section \ref{arch}. All of the aforementioned models can be found in \newline https://github.com/pomonam/LearnablePoolingMethods, with complete documentation and easily usable modules.

\subsection{Training Details}
The Youtube-8M dataset is split into training (70\%), validation (20\%) and testing (10\%). For the sake of accuracy, we utilize both training and validation portions of the Youtube-8M dataset for training, while leaving out a random 2\% of data for the sake of local validation testing.

All transformer related experiments are trained using the Adam optimizer \cite{adam} with an initial learning rate of 0.0003 and a batch size of 32 (64 in all) on two NVIDIA P100 GPUs. All function approximation related experiments are performed using a batch size of 4 (32 in all) on eight NVIDIA K80 GPUs. For each video, we utilized uniform sampling of 256 frames, to be able evenly select features consistently, while maintaining temporal consistency, for the sake of attention-based models.

We did not train many of our models exhaustively due to time and resource constraints while participating in this competition. We stop training early after roughly 3 epochs (270k steps). For complete implementation details, visit our GitHub repository.

\subsection{Testing Results}
In the "The 2nd YouTube-8M Video Understanding Challenge", models are evaluated using the Global Average Precision (GAP) metric. In Equation \eqref{gap}, p(i) and r(i) refer to the precision and recall of the top i predictions, respectively.

\begin{align} 
  GAP = \sum_{k=1}^{20} p(i) * r(i) \label{gap}
\end{align}

\setlength{\tabcolsep}{4pt}
\begin{table}
\begin{center}
\caption{A collection of the highest testing scores, as determined by Global Average Precision (\textbf{GAP}). Second Order refers to Section \ref{experimental}, Attention Enhanced refers to Section \ref{modelone} and Attention NetVLAD refers to Section \ref{modeltwo}}
\label{results}
\begin{tabular}{llll}
\hline\noalign{\smallskip}
Name & Training Steps & Batch Size & Public Test \textbf{GAP}\\
\hline
\noalign{\smallskip}
Baseline NetVLAD & 270k & 80 & \textbf{0.870} \\
Second Order & 270k & 32 & 0.865 \\
Attention Enhanced & 270k & 64 & 0.856 \\
Attention NetVLAD & 220k & 64 & \textbf{0.867} \\

\hline
\end{tabular}
\end{center}
\end{table}

Our results are encouraging, but currently, do not improve the state of the art. Our largest issue for the models listed in Table \ref{results}, as well as other models that exist in our GitHub repository, is the problem of generalization and/or overfitting, which is a relatively poorly understood topic. During training, we clearly have the capacity to learn training regularities, however, we inevitably overfit, even when experimenting with common techniques such as dropout and early stop.

Despite our misfortune, it is possible to extend our second order models by creating a parallelizable Sinkhorn scaling algorithm to make use of crucial democratic weights, or by avoiding the stage of projecting input features into a low dimensional space (given hardware resources for such a model). In addition, exploration of other regularization techniques, such as regularizer cost functions from Section \ref{experimental} are promising.

\section{Conclusions}
In this paper, we made modifications to the state-of-the-art approaches for aggregating local video descriptors. We drew a connection between NetVLAD and Transformers to learn cluster similarities per local descriptors. In addition, we had some encouraging results using a function approximation approach. These techniques increased model capacity compared to the state of the art. Experimental results demonstrate that the testing accuracy is similar to that of the state of the art. Due to the time constraints of the competition, we did not fully investigate parameter tuning or overfitting issues. We plan to explore regularization costs, as well as other avenues discussed for accuracy improvement.

%
%
%
%

\end{document}